\documentclass[letterpaper, 10 pt, conference]{ieeeconf}

\IEEEoverridecommandlockouts
\usepackage{amsmath}
\usepackage{amssymb}
\usepackage{algorithm}
\usepackage{algorithmic}
\usepackage{graphicx}
\usepackage{multirow}
\usepackage{xcolor}
\makeatletter
\let\NAT@parse\undefined
\makeatother
\usepackage[numbers,sort]{natbib}

\usepackage[breaklinks,colorlinks=true,citecolor=green]{hyperref}

\newcommand{\algcomment}[1]{\textcolor{gray}{// #1}}
\title{\LARGE \bf
FA-RDP: A Frequency-Adaptive Reactive Diffusion Policy for Contact-Rich Manipulation
}

\author{
Lifeng Zhuo\textsuperscript{*1} \quad
Wendi Chen\textsuperscript{*1,2} \quad
Han Xue\textsuperscript{1} \quad
Shirun Tang\textsuperscript{3} \quad
Jun Lv\textsuperscript{3} \\
Cewu Lu\textsuperscript{1,2,3\(\dagger\)} \quad
Chuan Wen\textsuperscript{1\(\dagger\)}%
\thanks{\textsuperscript{1}Shanghai Jiao Tong University \quad
\textsuperscript{2}Shanghai Innovation Institute \quad
\textsuperscript{3}Noematrix Ltd. \quad
\textsuperscript{*}Equal contribution \quad
\textsuperscript{\(\dagger\)}Corresponding authors.}
}

\begin{document}

\maketitle
\thispagestyle{empty}
\pagestyle{empty}
\raggedbottom

\begin{abstract}
In contact-rich manipulation, action multimodality and reactivity dominate
different stages of a single episode. Before contact, multiple trajectories
might be equally valid, making it important to preserve diverse action modes.
After contact, geometric constraints and force limits narrow the solution space,
while successful execution demands rapid responses to force feedback. However,
standard diffusion policies use a fixed inference frequency and sampling steps
throughout the episode, forcing a fundamental compromise: low-frequency,
multi-step sampling better preserves pre-contact multimodality but responds
slowly to force feedback, whereas high-frequency sampling improves reactivity
but tends to collapse distinct pre-contact modes. To resolve this tradeoff, we
present \textbf{FA-RDP}, a frequency-adaptive reactive diffusion policy. A shared
multi-frequency visual-force Transformer predicts action chunks at both low and
high frequencies, while a learned multimodality indicator dynamically selects
multi-step low-frequency sampling before contact and one-step high-frequency
sampling as action ambiguity decreases. We further introduce Manifold
Consistency Distillation (MCD), which reparameterizes the diffusion network to
predict actions on the robot action manifold while retaining DDPM-based residual
supervision. Experiments on three contact-rich manipulation tasks show that
FA-RDP achieves the highest success rate while preserving diverse pre-contact
trajectory modes. Code and demos are available at
\href{https://fa-rdp.github.io/}{fa-rdp.github.io}.
\end{abstract}

\section{Introduction}

\begin{figure}[t]
\centering
\includegraphics[width=\columnwidth]{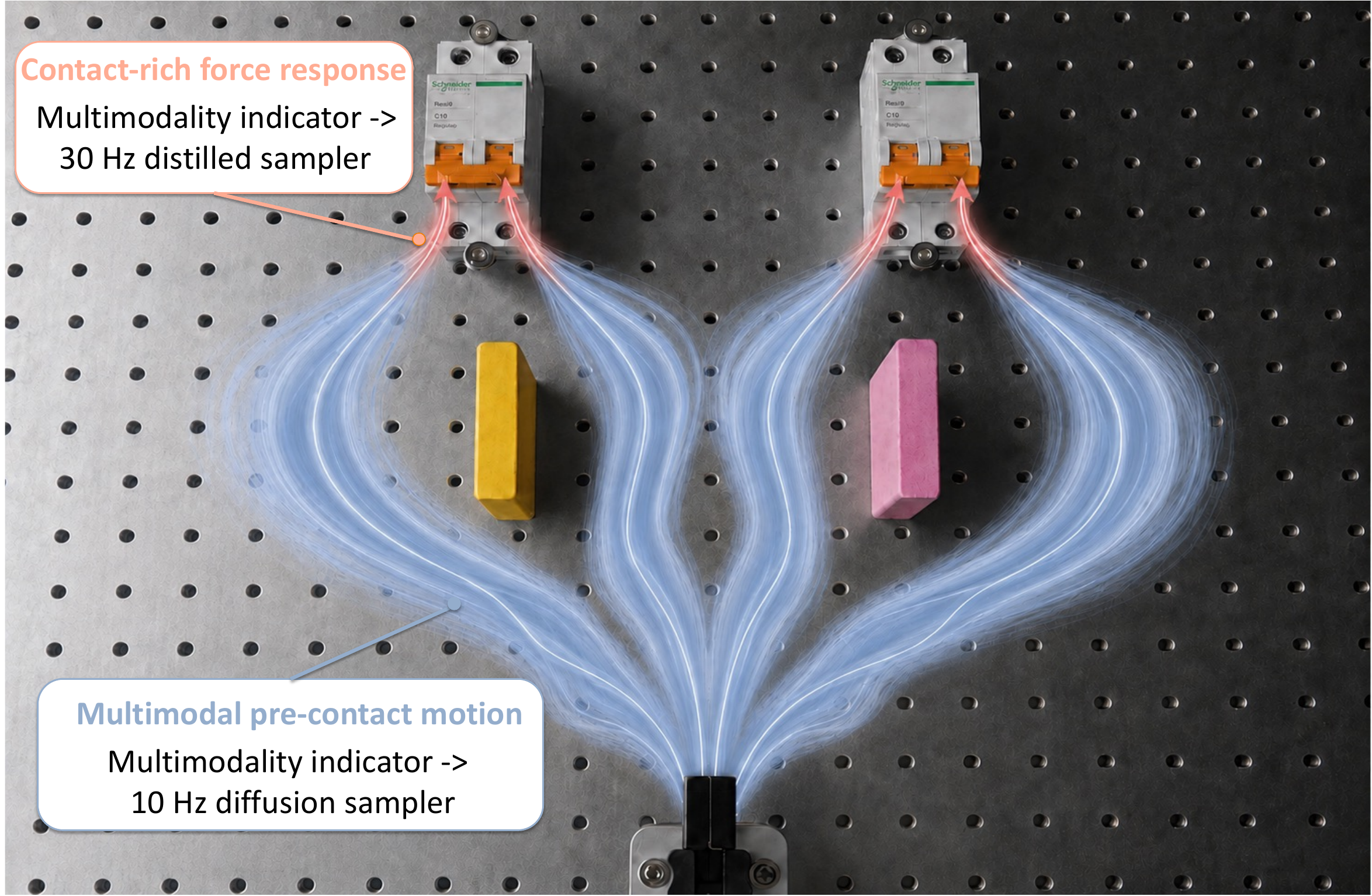}
\caption{Multimodality-guided frequency adaptation across manipulation phases.
Before contact, low indicator value reflects multiple possible approach
trajectories and selects the low-frequency diffusion sampler. After contact, the
indicator value rises as physical constraints reduce action ambiguity, selecting the
high-frequency distilled sampler for rapid force-feedback response.}
\label{fig:contact_phase_motivation}
\end{figure}

Contact-rich manipulation requires a robot policy to solve two phase-dependent
control problems: \textbf{diversity} before contact and \textbf{reactivity}
after contact \cite{tsuji2025survey}. As shown in
Fig.~\ref{fig:contact_phase_motivation},
pre-contact visual observations can admit several valid approach or contact
modes, requiring the policy to preserve diverse pre-contact trajectory modes. By contrast, once contact is established,
the robot must maintain contact and regulate the applied force under physical
constraints, requiring rapid action updates using measured wrench feedback.

Robot policies such as Diffusion Policy \cite{chi2025diffusion} predict
temporally coherent action sequences, but control within each chunk still 
remains open-loop. After contact, the policy
cannot use newly observed force signals to update actions, so
small pose errors can quickly cause force spikes, slip, or contact loss.
Closed-loop execution shortens this feedback delay by conditioning each executed
action on immediate force observations, maintaining contact before errors
accumulate. Previous works like RDP \cite{xue2025reactive} realize closed-loop execution
using a hierarchical slow-fast design, but information can be lost between its slow
and fast policies because the fast policy relies only on compressed latent actions.
ImplicitRDP \cite{chen2026implicitrdp} removes this explicit hierarchy by learning
visual and force modalities in an end-to-end diffusion network. However, the
multi-step diffusion sampling limits the closed-loop inference frequency. These
limitations motivate our central question:
Can an end-to-end visual-force diffusion policy achieve high-frequency
closed-loop response while preserving pre-contact multimodality?

To raise the closed-loop frequency, the most direct approach is to reduce the
number of denoising steps through distillation
\cite{salimans2022progressive,prasad2024consistency}, which improves reactivity.
But applying a one-step sampler throughout the episode sacrifices multimodality:
it can collapse the multimodal action distribution needed before contact and
degrade action smoothness, which may break contact or violate force constraints.
The policy must therefore preserve multimodal sampling before contact while
producing smooth actions at a high control rate after contact.
This calls for adapting the inference frequency to the manipulation phase,
rather than fixing it throughout the episode.

We propose FA-RDP, a frequency-adaptive reactive visual-force diffusion policy
for contact-rich manipulation, built from three components. First, a single
visual-force diffusion backbone is shared across frequencies through
frequency-aware positional encoding, so the same network predicts both
low-frequency and high-frequency action chunks without training separate models.
Second, a learned multimodality indicator, computed from the visual input,
estimates how multimodal the current pre-contact behavior is, and selects the
low-frequency multi-step sampler while this ambiguity is high and the
high-frequency distilled sampler once contact constraints reduce it. Third,
because the selected high-frequency sampler would still require multiple
denoising steps, we distill it with manifold consistency distillation for
one-step inference.

Experiment results show that FA-RDP achieves the highest average success rate of
$81.7\%$, 30.0 percentage points higher than baseline. The component analyses further support the phase structure in
Fig.~\ref{fig:contact_phase_motivation}: FA-RDP preserves diverse pre-contact
modes, while a higher indicator value in lower-multimodality contact-rich phases
activates the distilled high-frequency sampler.

The main contributions of this work are:
\begin{itemize}
    \item We propose FA-RDP, a frequency-adaptive reactive visual-force
    diffusion policy, which is guided by a multimodality indicator and preserves
    pre-contact diversity while enabling post-contact reactivity.
    \item We introduce manifold consistency distillation, which reparameterizes
    the diffusion network to predict action chunks on the robot action manifold,
    rather than noise-like epsilon, score, or velocity targets, while retaining
    DDPM-based residual supervision for stable one-step high-frequency inference.
    \item We evaluate FA-RDP on three contact-rich manipulation tasks,
    showing improved task success, preserved pre-contact multimodality, and the
    benefit of higher-rate closed-loop control in contact-rich tasks.
\end{itemize}

\section{Related Work}
\label{sec:related_work}

\subsection{Robot Learning with Force Input}

Vision-only policies such as Diffusion Policy \cite{chi2025diffusion}, ACT
\cite{zhao2023learning}, and 3D Diffusion Policy \cite{ze20243d} do not use
immediate force feedback, which limits their reactivity during contact.

To address this limitation, robot learning methods increasingly incorporate
force or tactile input for contact-rich manipulation. DexForce \cite{chen2025dexforce}
and 3D-ViTac \cite{huang20243d} condition action prediction on force or contact
signals; other works \cite{liu2025factr,oh2026factr,yu2026forcevla,guzey2023dexterity}
learn force-aware representations; and recent works
\cite{yuan2026vtam,yuan2026ftp} scale tactile and video-tactile data toward
generalist policies. Collectively, these methods show that contact signals carry
information unavailable from vision, but action execution within each predicted
chunk still remains open-loop, limiting real-time reaction to force feedback.

Previous works implement high-frequency force reaction in different ways: Compliant
Residual DAgger \cite{xu2026compliant} and Force Policy \cite{fang2026force} rely on a
separate low-level force controller, RDP \cite{xue2025reactive} uses a hierarchical
slow-fast design but relies on a compressed latent interface, and ImplicitRDP
\cite{chen2026implicitrdp} learns visual and force modalities end-to-end but still
requires multi-step denoising. FA-RDP
instead keeps an end-to-end visual-force policy and makes the inference
frequency adaptive.

\subsection{Multi-Frequency Policy Learning}

Inference frequency is a key design choice for robot policies.
Low-frequency inference supports longer-horizon and multimodal behavior, whereas
high-frequency inference improves force reactivity but increases sampling cost
and training difficulty. Prior work studies this tradeoff through different
mechanisms: HiPolicy \cite{zhang2026hipolicy} uses hierarchical multi-frequency action
chunking, high-frequency latent chunk learning \cite{wang2026learning}
moves dense actions into a latent space, DVAC \cite{feng2026denoising} adapts replanning from
denoising variance, and AHA-WAM \cite{cai2026aha} separates low-frequency world
planning from high-frequency action execution. ManipForce
\cite{lee2025manipforce} uses frequency-aware representations to fuse
asynchronous visual and force observations.

Existing multi-frequency policies are primarily motivated by hierarchical
control, latent action learning, adaptive replanning, or asynchronous sensing.
FA-RDP is instead motivated by the phase-dependent
requirements of contact-rich manipulation: multimodal pre-contact behavior and
rapid force-feedback response after contact. Accordingly, it uses
frequency-aware positional encoding in a shared visual-force diffusion backbone
to predict low-frequency and high-frequency action chunks.

\subsection{Robot Policy Distillation}

Predicting high-frequency action chunks alone is not sufficient: closed-loop
execution also requires the sampler to be fast enough for real-time control.
DDIM \cite{song2020denoising}
provides deterministic denoising trajectories for diffusion models, and
progressive distillation \cite{salimans2022progressive} reduces the number of
sampling steps by training a student sampler to match longer teacher
trajectories. Consistency Models \cite{song2023consistency} learn mappings that
support one-step or few-step generation. In robotics, ManiCM
\cite{lu2024manicm}, Consistency Policy \cite{prasad2024consistency}, and Hybrid
Consistency Policy \cite{zhao2026hybrid} adapt consistency-based
acceleration to visuomotor policies. Flow-based policies pursue the
same low-latency goal from another parameterization: MeanFlow \cite{geng2026mean}
learns average velocity fields for one-step generation, Mean Flow Policy
\cite{sheng2026mp1} applies mean-velocity objectives to robot
manipulation, and FreqPolicy \cite{su2026freqpolicy} improves flow-based visuomotor
policies with frequency consistency.

These methods demonstrate effective one-step or few-step inference, but
epsilon, score, or velocity targets are difficult to distill for robot
action prediction. Inspired by JiT \cite{li2026back} and Pixel Mean
Flows \cite{lu2026one}, FA-RDP instead predicts action chunks on the robot
action manifold.

\section{Preliminary}
\label{sec:preliminaries}

We adopt the consistent reactive diffusion inference mechanism from prior
end-to-end visual-force diffusion policies \cite{chen2026implicitrdp}. This mechanism
combines slow visual-proprioceptive tokens and fast force tokens in one
Transformer. Its structural slow-fast learning separates observations into a
slow part for visual context and a fast part for dense force feedback. During
training, a causal action-force mask ensures that each action token can attend
only to current and past force tokens, preventing future contact leakage while
keeping parallel diffusion training.

At inference, the consistent mechanism realizes closed-loop force control
within an action chunk. DDIM with $\eta=0$ makes the denoising trajectory
deterministic for a fixed initial noise $\mathbf{A}_K$. The policy therefore
caches the slow context and $\mathbf{A}_K$ at the beginning of a chunk, then
refreshes fast force tokens at each control step, runs multi-step denoising with
the cached context and updated force tokens, and executes the newest predicted
action. However, running multi-step denoising at each step limits the
closed-loop force-feedback frequency, restricting rapid correction during
continuous contact-rich interaction.

\section{Method}
\label{sec:method}

FA-RDP addresses the low force-feedback frequency of prior end-to-end policies
with a multi-frequency visual-force diffusion backbone. It preserves multimodal pre-contact behavior
while enabling high-frequency control
through three components: a multi-frequency visual-force
Transformer that supports multi-frequency action prediction
(Sec.~\ref{subsec:multi_frequency_transformer}), multimodality-based frequency
selection that chooses the appropriate sampling mode
(Sec.~\ref{subsec:multimodality}), and manifold consistency distillation (MCD) that
stabilizes one-step high-frequency action prediction
(Sec.~\ref{subsec:distillation}). Implementation details are provided in
Sec.~\ref{subsec:implementation_details}.

\subsection{Multi-Frequency Visual-Force Transformer}
\label{subsec:multi_frequency_transformer}

FA-RDP realizes frequency-adaptive visual-force control through a shared
multi-frequency visual-force Transformer. The same backbone is used for both
frequency modes. A single forward pass is conditioned on one
frequency mode $\nu\in\{\ell,h\}$ and predicts the action chunk for that mode.
This shared backbone is trained in Stage 1 with joint low-frequency and
high-frequency diffusion objectives.

\begin{figure}[t]
\centering
\includegraphics[width=\columnwidth]{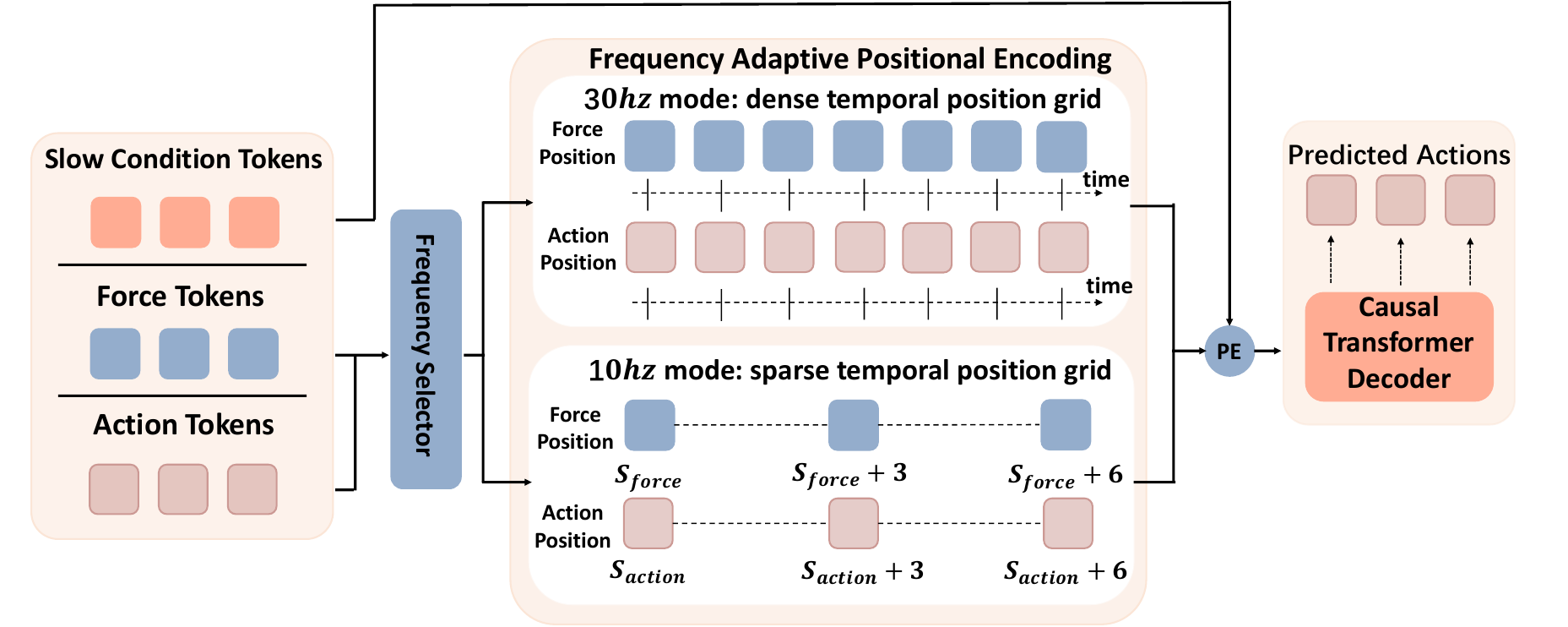}
\caption{Multi-frequency visual-force Transformer. The frequency mode selects
frequency-adaptive positional indices on a shared temporal grid: dense
positions for high-frequency action chunks and sparse positions for
low-frequency action chunks. The causal decoder supports closed-loop updates by
preventing action tokens from accessing future action or force tokens.}
\label{fig:multi_frequency_transformer}
\end{figure}

The key design is frequency-adaptive positional encoding, as shown in
Fig.~\ref{fig:multi_frequency_transformer}. The frequency mode is not embedded
as an extra token; instead, it selects temporal indices on a shared
temporal grid. The high-frequency mode uses consecutive positions, whereas
the low-frequency mode uses sparse positions, so both modes are expressed on
the same underlying clock. This design lets the backbone share weights
across frequencies while preserving each mode's temporal layout. We use 
causal action-force mask to avoid future-contact leakage.

For each mode $\nu$, let $\mathbf{A}^\nu$ denote the normalized demonstration action
sequence for sample $n$, conditioned on slow observation $\mathbf{o}_n^s$
and force observation $\mathbf{o}_n^{\mathrm{force}}$. The forward diffusion
process samples a timestep $k$ and noise
$\boldsymbol{\epsilon}\sim\mathcal{N}(0,\mathbf{I})$, then constructs
\begin{equation}
    \mathbf{A}^\nu_k =
    \sqrt{\bar{\alpha}_k} \mathbf{A}^\nu +
    \sqrt{1-\bar{\alpha}_k}\boldsymbol{\epsilon} .
    \label{eq:forward_diffusion}
\end{equation}
Diffusion timesteps run from $0$ to $K$, where $K$ is the initial-noise
timestep; we omit the subscript $0$ for clean demonstration and predicted
action chunks.
Throughout, $\mathbf{A}$ denotes a given or constructed action state, whereas
$\hat{\mathbf{A}}$ denotes an action state predicted by a network.
We parameterize the network to predict the action chunk
$\hat{\mathbf{A}}^\nu=\pi_\theta(\mathbf{A}^\nu_k,k,
\mathbf{o}_n^s,\mathbf{o}_n^{\mathrm{force}},\nu)$. The
implied noise prediction is
\begin{equation}
    \hat{\boldsymbol{\epsilon}} =
    \frac{\mathbf{A}^\nu_k-\sqrt{\bar{\alpha}_k}\hat{\mathbf{A}}^\nu}
    {\sqrt{1-\bar{\alpha}_k}},
    \label{eq:eps_from_sample}
\end{equation}
and we optimize the per-mode diffusion objective
\begin{equation}
    \mathcal{L}_{\epsilon}^{\nu} =
    \mathbb{E}_{\mathbf{A}^\nu,k,\boldsymbol{\epsilon}}
    \left[\|\hat{\boldsymbol{\epsilon}}-\boldsymbol{\epsilon}\|_2^2\right].
    \label{eq:eps_loss}
\end{equation}
The shared backbone is trained with a joint multi-frequency diffusion objective
that applies the same loss to the low-frequency and high-frequency targets,
\begin{equation}
    \mathcal{L}_{\mathrm{MF}} =
    \mathcal{L}_{\epsilon}^{\ell} + \mathcal{L}_{\epsilon}^{h}.
    \label{eq:multifreq_loss}
\end{equation}
Thus, the network predicts action samples, while the training objective still
uses epsilon supervision.

\subsection{Multimodality-Based Frequency Selection}
\label{subsec:multimodality}

The multi-frequency Transformer provides both low-frequency and high-frequency diffusion
samplers, but the policy still needs to decide which sampler to use at different
phases. We use a learned indicator that estimates local multimodality instead of relying on a phase label.

Stage 2 trains the multimodality indicator head on an independent calibration
set. The Stage 1 visual-force policy is frozen, so this training does not change
the multi-frequency backbone. For each calibration example, we fix the same
visual-force condition and demonstration target, independently sample the
low-frequency policy $N=8$ times with different initial diffusion noise, and
compute an empirical action residual:
\begin{equation}
    e(\mathbf{o}_n)=
    \frac{1}{N}
    \sum_{m=1}^{N}
    \left\|\hat{\mathbf{A}}^{\ell}_m-\mathbf{A}^{\ell}\right\|_2^2 .
    \label{eq:multimodality_error}
\end{equation}
Here, $\hat{\mathbf{A}}^{\ell}_m$ is the $m$-th action chunk sampled from the
frozen Stage 1 low-frequency policy, and $\mathbf{A}^\ell$ is the corresponding
demonstration.
A Transformer indicator head maps slow observation tokens to a scalar logit
$s(\mathbf{o}_n)$, with
$\sigma(\mathbf{o}_n)=1+\exp(s(\mathbf{o}_n))$. Following the
confidence-weighted regression form of VGGT \cite{wang2025vggt}, we train the indicator
with
\begin{equation}
    \mathcal{L}_{\mathrm{MM}} =
    \mathbb{E}_{\mathbf{o}_n}
    \left[\gamma e(\mathbf{o}_n)\sigma(\mathbf{o}_n)
    - \alpha_{\mathrm{MM}}\log\sigma(\mathbf{o}_n)\right],
    \label{eq:multimodality_loss}
\end{equation}
where $\gamma$ and $\alpha_{\mathrm{MM}}$ control the error penalty and
regularizer.

At inference time, FA-RDP computes the indicator value from the current slow tokens
and selects the sampler with a fixed threshold. This makes execution
frequency-adaptive, but the selected high-frequency diffusion sampler still
requires multi-step denoising. We therefore distill the high-frequency sampler
for fast force-feedback response.

\subsection{Manifold Consistency Distillation for One-Step Control}
\label{subsec:distillation}

\begin{figure}[t]
\centering
\includegraphics[width=\columnwidth]{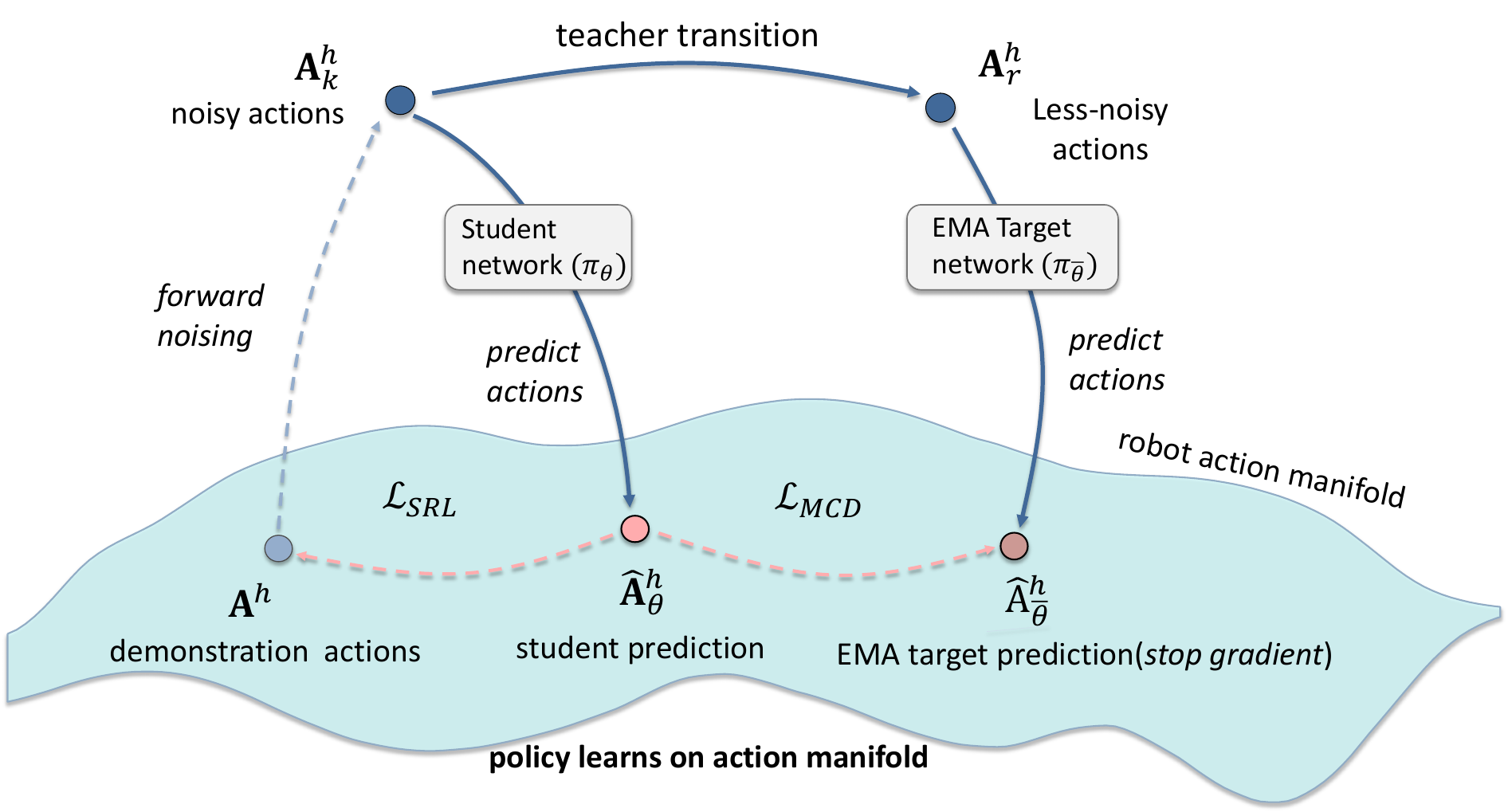}
\caption{Manifold consistency distillation for the high-frequency sampler. A
teacher transition maps $\mathbf{A}^{h}_k$ to a less noisy state $\mathbf{A}^{h}_r$.
Both networks predict action chunks on the robot action manifold. MCD aligns the
student prediction with the stop-gradient target prediction, while SRL
regularizes the student prediction to stay close to the demonstration action.}
\label{fig:mcd_distillation}
\end{figure}

Manifold consistency distillation (MCD) turns the selected high-frequency sampler into
a one-step sampler. Directly distilling a model that predicts epsilon, score, or
velocity requires the student to fit high-frequency, noise-like targets, which
makes stable action prediction difficult. Inspired by the sample-prediction view
of JiT \cite{li2026back} and Pixel Mean Flows \cite{lu2026one}, we reparameterize the
denoising model to predict action chunks on the robot action manifold. The fixed
DDPM conversion in Eq.~\ref{eq:eps_from_sample} retains residual supervision
without making the network predict a noise-like target, reducing the student's
learning difficulty.
Fig.~\ref{fig:mcd_distillation} illustrates the resulting distillation process.

The consistency objective follows the general goal of reducing diffusion NFEs
via progressive distillation \cite{salimans2022progressive} and is related to
Consistency Models \cite{song2023consistency} and Consistency Policy
\cite{prasad2024consistency}. Under the shared condition $\mathbf{c}_n^h$, let
$\hat{\mathbf{A}}_{\theta}^h\equiv
\pi_\theta(\mathbf{A}_k^h,k,\mathbf{c}_n^h)$ and
$\hat{\mathbf{A}}_{\bar{\theta}}^h\equiv
\pi_{\bar{\theta}}(\mathbf{A}_r^h,r,\mathbf{c}_n^h)$ denote the two action
predictions. In Stage 3, the student is trained with
\begin{equation}
\begin{aligned}
    \mathcal{L}_{\mathrm{distill}} =
    &\lambda_{\mathrm{MCD}}
    \left\|\hat{\mathbf{A}}_{\theta}^h
    - \operatorname{sg}[\hat{\mathbf{A}}_{\bar{\theta}}^h]\right\|_2^2 \\
    &+ \lambda_{\mathrm{SRL}}
    \left\|\hat{\mathbf{A}}_{\theta}^h - \mathbf{A}^h\right\|_2^2 .
\end{aligned}
    \label{eq:distill}
\end{equation}
Here, $k>r$ are adjacent grid timesteps for the noisy state
$\mathbf{A}_k^h$ and less-noisy teacher state $\mathbf{A}_r^h$, conditioned on
$\mathbf{c}_n^h=(\mathbf{o}_n^s,\mathbf{o}_n^{\mathrm{force}},h)$.
$\theta$ and $\bar{\theta}$ are the student and EMA target parameters,
$\operatorname{sg}[\cdot]$ stops gradients through the target prediction, and
$\mathbf{A}^h$ is the demonstrated high-frequency action chunk.
$\lambda_{\mathrm{MCD}}$ and $\lambda_{\mathrm{SRL}}$ weight the MCD and
sample regression loss (SRL) terms. After distillation, FA-RDP uses the
multi-frequency checkpoint for the low-frequency sampler and the distilled
checkpoint for the high-frequency sampler. The final inference procedure follows
ImplicitRDP's consistent inference mechanism \cite{chen2026implicitrdp} and is summarized
in Alg.~\ref{alg:multimodality_inference}: cache slow context and initial noise at
chunk start, select the frequency using the indicator, refresh wrench tokens at each
step, run multi-step DDIM for the low-frequency mode or direct one-step
action prediction for the high-frequency mode, and execute the newest valid action.

\begin{algorithm}[t]
\caption{FA-RDP Consistent Frequency-Adaptive Inference}
\label{alg:multimodality_inference}
\footnotesize
\begin{algorithmic}[1]
\REQUIRE Policy networks $\pi_{\theta_\ell}$ and $\pi_{\theta_h}$,
slow observation encoder $\mathcal{E}_{\mathrm{slow}}$, fast observation
encoders $\mathcal{E}_{\mathrm{fast}}^\ell$ and
$\mathcal{E}_{\mathrm{fast}}^h$, slow observation horizon $h_o$,
execution horizons $h_e^\ell$ and $h_e^h$, latency steps $l^\ell$ and $l^h$,
multimodality indicator $\sigma$, threshold $\tau$
\STATE Initialize step $t \leftarrow 0$
\STATE \algcomment{slow loop: action chunk modeling}
\WHILE{Task not done}
    \STATE \algcomment{cache slow context and noise and select frequency}
    \STATE $\mathbf{O}_{\mathrm{slow}} \leftarrow
    \mathrm{GetSlowObservation}(\mathrm{len}=h_o)$
    \STATE $\mathbf{Z}_{\mathrm{slow}} \leftarrow
    \mathcal{E}_{\mathrm{slow}}(\mathbf{O}_{\mathrm{slow}})$
    \STATE $q \leftarrow \sigma(\mathbf{Z}_{\mathrm{slow}})$
    \IF{$q > \tau$}
        \STATE $\nu \leftarrow h$, $h_e \leftarrow h_e^h$, $l \leftarrow l^h$
        \STATE $\pi \leftarrow \pi_{\theta_h}$,
        $\mathcal{E}_{\mathrm{fast}} \leftarrow \mathcal{E}_{\mathrm{fast}}^h$
    \ELSE
        \STATE $\nu \leftarrow \ell$, $h_e \leftarrow h_e^\ell$, $l \leftarrow l^\ell$
        \STATE $\pi \leftarrow \pi_{\theta_\ell}$,
        $\mathcal{E}_{\mathrm{fast}} \leftarrow \mathcal{E}_{\mathrm{fast}}^\ell$
    \ENDIF
    \STATE $\mathbf{A}_K^\nu \leftarrow \mathcal{N}(0,\mathbf{I})$
    \STATE \algcomment{fast loop: closed-loop control within the horizon}
    \FOR{$i \leftarrow 0$ to $h_e-1$}
        \STATE \algcomment{update fast context}
        \STATE $\mathbf{O}_{\mathrm{fast}} \leftarrow
        \mathrm{GetFastObservation}(\nu,\mathrm{len}=h_o+i+l)$
        \STATE $\mathbf{Z}_{\mathrm{fast}} \leftarrow
        \mathcal{E}_{\mathrm{fast}}(\mathbf{O}_{\mathrm{fast}})$
        \STATE \algcomment{get a noisy action sequence of a certain length}
        \STATE $\mathbf{A}_{\mathrm{pre}}^\nu \leftarrow
        \mathrm{Prefix}(\mathbf{A}_K^\nu,i+h_o+l)$
        \STATE \algcomment{consistent denoising with cache}
        \IF{$\nu = h$}
            \STATE $\hat{\mathbf{A}}^\nu \leftarrow
            \pi(\mathbf{A}_{\mathrm{pre}}^\nu,K,\mathbf{Z}_{\mathrm{slow}},
            \mathbf{Z}_{\mathrm{fast}},\nu=h)$
        \ELSE
            \STATE $\hat{\mathbf{A}}^\nu \leftarrow
            \mathrm{DDIM}(\pi,\mathbf{Z}_{\mathrm{slow}},
            \mathbf{Z}_{\mathrm{fast}},\mathbf{A}_{\mathrm{pre}}^\nu,
            \nu=\ell,\eta=0)$
        \ENDIF
        \STATE \algcomment{execute the current step action}
        \STATE $\mathbf{a}_t \leftarrow \hat{\mathbf{A}}^\nu[-1]$
        \STATE Execute $\mathbf{a}_t$
        \STATE $t \leftarrow t + 1$
        \IF{Task done}
            \STATE \textbf{break}
        \ENDIF
    \ENDFOR
\ENDWHILE
\end{algorithmic}
\end{algorithm}

\subsection{Implementation Details}
\label{subsec:implementation_details}

\textbf{100 Hz force compensation.}
Our demonstrations record robot states rather than commanded targets. During
deployment, impedance control can therefore create a tracking gap between a
policy-predicted state and the robot's realized motion. To compensate for this
execution error, all methods share a 100 Hz command layer that linearly
interpolates policy outputs and adjusts each command using the latest measured
world-frame external force:
\begin{equation}
    \mathbf{p}_{\mathrm{cmd}} =
    \mathbf{p}_{\mathrm{policy}} - \lambda \mathbf{f}_{\mathrm{ext}},
    \quad \lambda = 10^{-4}.
    \label{eq:force_compensation}
\end{equation}
This translation-only compensation is applied before sending targets through the
Flexiv Cartesian motion-force interface; unlike ImplicitRDP, which
additionally finetunes the low-level controller for precise position tracking,
all methods here share the same default control layer; this controller-level
difference causes the performance gap.

\textbf{Algorithm settings.}
Frequency-adaptive execution requires the low-frequency and high-frequency modes to span
the same prediction horizon at different temporal resolutions. We therefore use a
low-frequency mode at 10 Hz with $H_{\ell}=16$ and a high-frequency mode at
30 Hz with $H_h=48$, so both modes cover the same 1.6 s horizon with different
action densities. Camera observations are recorded at 10 Hz, and the
force/torque stream used by the policy is recorded at 30 Hz. During execution, the slow
loop refreshes every 1 s, corresponding to $h_e^\ell=10$ for the low-frequency
mode and $h_e^h=30$ for the high-frequency mode. The diffusion process uses 100
training timesteps, indexed from 0 to 99. To make the high-frequency mode fast
enough for closed-loop use, we distill it on the six-timestep grid
$\mathcal{G}=\{99,79,59,39,19,0\}$.

\textbf{Hardware settings.}
We run policy inference on an Intel Core Ultra 9 285K CPU and NVIDIA GeForce
RTX 5090 GPU. The policy inference latency is below 30 ms for high-frequency
execution and below 50 ms for low-frequency execution.

\section{Experiments}
\label{sec:experiments}

We evaluate FA-RDP on three real-world contact-rich manipulation tasks to answer four
key questions:
\begin{itemize}
    \item \textbf{Q1:} How does FA-RDP compare against visual-only diffusion
    policy, hierarchical visual-force policy, fixed-frequency end-to-end
    visual-force policy, and a non-diffusion regression baseline?
    \item \textbf{Q2:} Does FA-RDP preserve multimodal ability?
    \item \textbf{Q3:} Does the multimodality indicator improve performance by
    switching between frequency modes?
    \item \textbf{Q4:} Does our distillation perform better than other
    distillation methods?
\end{itemize}

\subsection{Experimental Setup}
\label{subsec:experiment_setup}

\begin{figure}[t]
\centering
\includegraphics[width=\columnwidth]{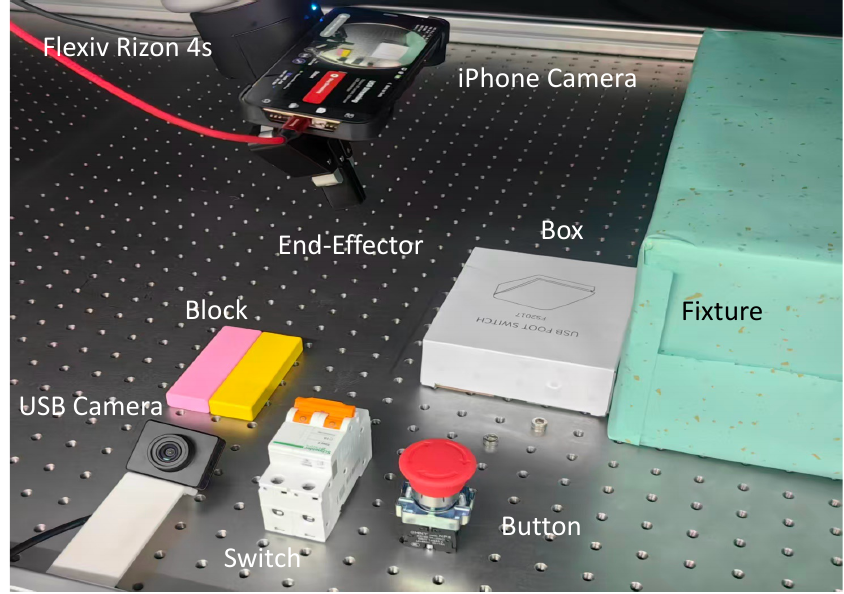}
\caption{Robot workspace and task objects. The Rizon 4s
follower arm executes tasks using observations from a wrist-mounted iPhone
camera and a fixed third-view USB camera, while the workspace contains the
target box, switch, button, front blocks, and fixture.}
\label{fig:experimental_setup}
\end{figure}

Our hardware setup uses a Flexiv Rizon 4R leader arm for teleoperation and a
Rizon 4s follower arm for task execution. We collect kinematic demonstrations
through TDK-enabled teleoperation, record contact forces from the follower's
end-effector F/T sensor, and capture visual observations with a wrist-mounted
iPhone fisheye camera~\cite{xue2026rethinking} and a fixed third-view USB camera.
Fig.~\ref{fig:experimental_setup} shows the hardware setup and task objects
used in our experiments.

We design three contact-rich manipulation tasks to evaluate multimodality and reactivity jointly.
In each task, two front blocks create multiple valid approaches, requiring the policy to choose a
collision-free trajectory before contact and then apply reactive control during
a force-sensitive flip, toggle, or press:
\begin{enumerate}
    \item \textbf{Dual Box Flipping:} The robot chooses a valid side approach
    and pushes a box against the fixture while maintaining contact.
    \item \textbf{Dual Switch Toggling:} The robot reaches the correct switch
    contact location, maintains directional contact along the switch motion,
    and applies sufficient force to complete the toggle.
    \item \textbf{Dual Button Pressing:} The robot aligns with the button axis
    at a nearly fixed contact point and applies force along the button axis
    until the actuation threshold is reached, with little tangential motion.
\end{enumerate}
For each task, we collect 60 demonstrations and evaluate 20 trials per method.

\subsection{Baselines}
\label{subsec:baselines}

We compare FA-RDP against the following main baselines in Q1:
\begin{itemize}
    \item \textbf{Diffusion Policy (DP):} vision-only CNN diffusion policy with
    open-loop action chunks.
    \item \textbf{Reactive Diffusion Policy (RDP):} hierarchical slow-fast
    visual-force policy.
    \item \textbf{ImplicitRDP:} fixed-frequency end-to-end visual-force
    diffusion policy.
    \item \textbf{Regression Policy w/ Force:} non-diffusion visual-force
    regression policy that predicts high-frequency action chunks using the same
    backbone as ImplicitRDP and an MSE objective.
    \item \textbf{FA-RDP (Ours):} multimodality-guided selection between the
    low-frequency diffusion sampler and high-frequency distilled sampler.
\end{itemize}

For Q2--Q4, we further evaluate sampler and distillation variants:
\begin{itemize}
    \item \textbf{High-frequency alone policy:} high-frequency diffusion
    policy without MCD or multimodality-guided selection.
    \item \textbf{High-frequency distilled alone policy:} high-frequency
    manifold-distilled policy without multimodality-guided selection.
    \item \textbf{MeanFlow Policy:} one-step policy whose backbone predicts
    mean and instantaneous velocity fields and updates the noisy trajectory with
    the mean velocity to predict an action chunk \cite{sheng2026mp1}.
    \item \textbf{Consistency Policy:} one-step policy that uses consistency
    distillation and EDM preconditioning to predict an action chunk from the raw
    network output \cite{prasad2024consistency}.
\end{itemize}

\subsection{Results and Analysis}
\label{subsec:results_analysis}

\subsubsection{Comparison with Baselines (Q1)}
As illustrated in Table~\ref{tab:main_results}, FA-RDP achieves the highest
success rate on all three tasks compared with visual-only diffusion,
hierarchical visual-force, fixed-frequency end-to-end visual-force, and direct
visual-force regression baselines.

\begin{table}[H]
\caption{Main success rates. Box, Button, and Switch denote the three
contact-rich tasks.}
\label{tab:main_results}
\centering
\footnotesize
\setlength{\tabcolsep}{2pt}
\begin{tabular*}{\columnwidth}{@{\extracolsep{\fill}}lcccc@{}}
\hline
\textbf{Method} & \textbf{Box} & \textbf{Button} & \textbf{Switch} & \textbf{Avg.} \\
\hline
DP & 0/20 & 2/20 & 4/20 & 10.0\% \\
RDP & 5/20 & 7/20 & 9/20 & 35.0\% \\
ImplicitRDP & 8/20 & 11/20 & 12/20 & 51.7\% \\
Regression Policy w/ Force & 2/20 & 4/20 & 6/20 & 20.0\% \\
\textbf{FA-RDP (Ours)} & \textbf{14/20} & \textbf{18/20} & \textbf{17/20} & \textbf{81.7\%} \\
\hline
\end{tabular*}
\end{table}

\begin{figure*}[t]
\centering
\includegraphics[width=\textwidth]{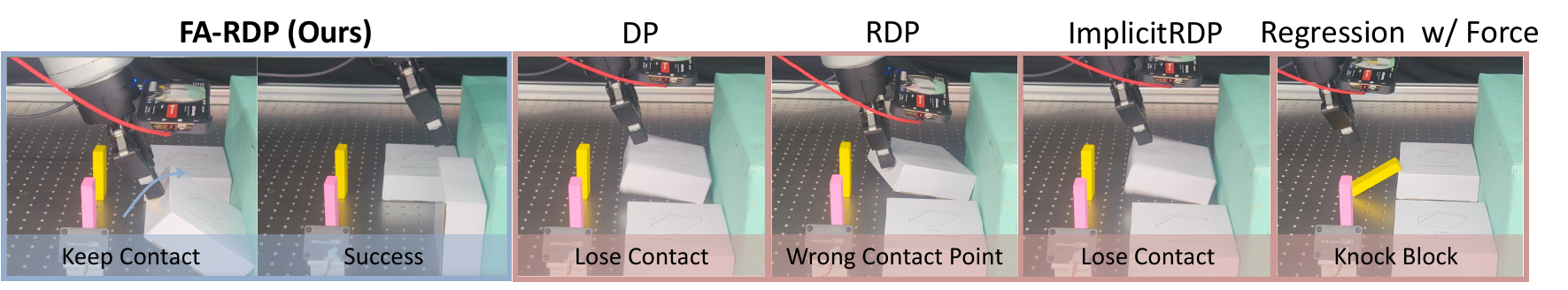}
\caption{Dual Box Flipping failure-case comparison. DP and ImplicitRDP can lose
contact during the flip, RDP can contact the wrong box region, and Regression
Policy w/ Force can knock down a front block before completing the task.
FA-RDP keeps contact and completes the box flipping task.}
\label{fig:dual_box_failure_cases}
\end{figure*}

\begin{figure*}[t]
\centering
\includegraphics[width=\textwidth]{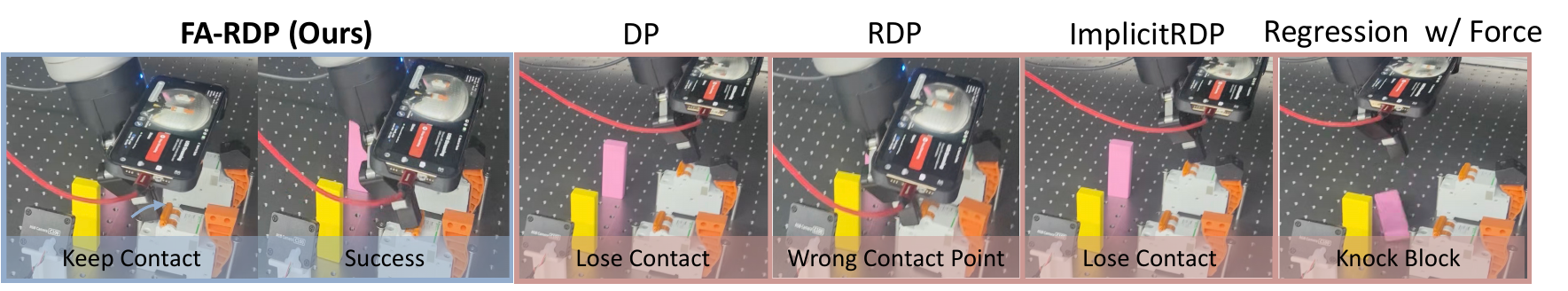}
\caption{Dual Switch Toggling failure-case comparison. DP and ImplicitRDP can
lose contact during the toggle, RDP can contact the wrong switch region, and
Regression Policy w/ Force can knock down a front block before completing the
task. FA-RDP keeps contact and completes the switch toggling task.}
\label{fig:dual_switch_failure_cases}
\end{figure*}

\begin{figure*}[t]
\centering
\includegraphics[width=\textwidth]{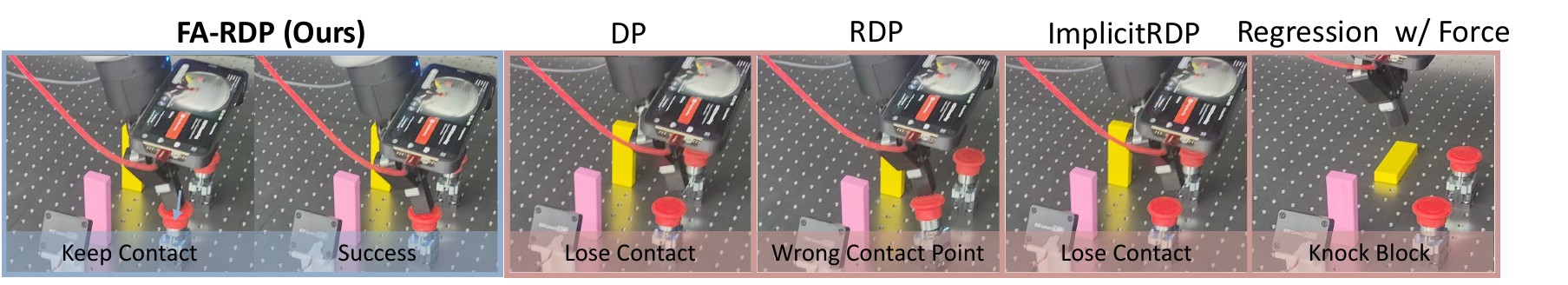}
\caption{Dual Button Pressing failure-case comparison. DP and ImplicitRDP can
lose contact during the press, RDP can contact the wrong button region, and
Regression Policy w/ Force can knock down a front block before completing the
task. FA-RDP keeps contact and completes the button pressing task.}
\label{fig:dual_button_failure_cases}
\end{figure*}

The failure cases in Figs.~\ref{fig:dual_box_failure_cases}--\ref{fig:dual_button_failure_cases}
reveal where each baseline breaks down. DP and ImplicitRDP both lose contact in
all three tasks, separating from the box during flipping, leaving the switch
before the toggle completes, and slipping off the button before the press
finishes. DP fails because its visual-only open-loop chunks cannot use new force
feedback, while ImplicitRDP predicts visual-force actions end to end but updates
too slowly to keep pace with the changing wrench after contact.

RDP reaches the task objects but contacts the wrong location, pushing the wrong
box region, reaching an incorrect switch point, and pressing away from the button
axis. These failures are consistent with spatial information loss through its
compressed slow-policy interface during precise approach.

Regression Policy w/ Force averages the multiple valid pre-contact approaches
under its MSE objective and cannot commit to a single collision-free mode, so it
knocks down the front block during target approach.

In contrast, FA-RDP separates the two phases: it keeps the multimodal
low-frequency sampler for a collision-free approach and switches to the distilled
high-frequency sampler after contact to react quickly to force feedback. This
preserves pre-contact diversity while maintaining contact through the flip,
toggle, and press.

\begin{figure}[H]
    \centering
    \includegraphics[width=\columnwidth]{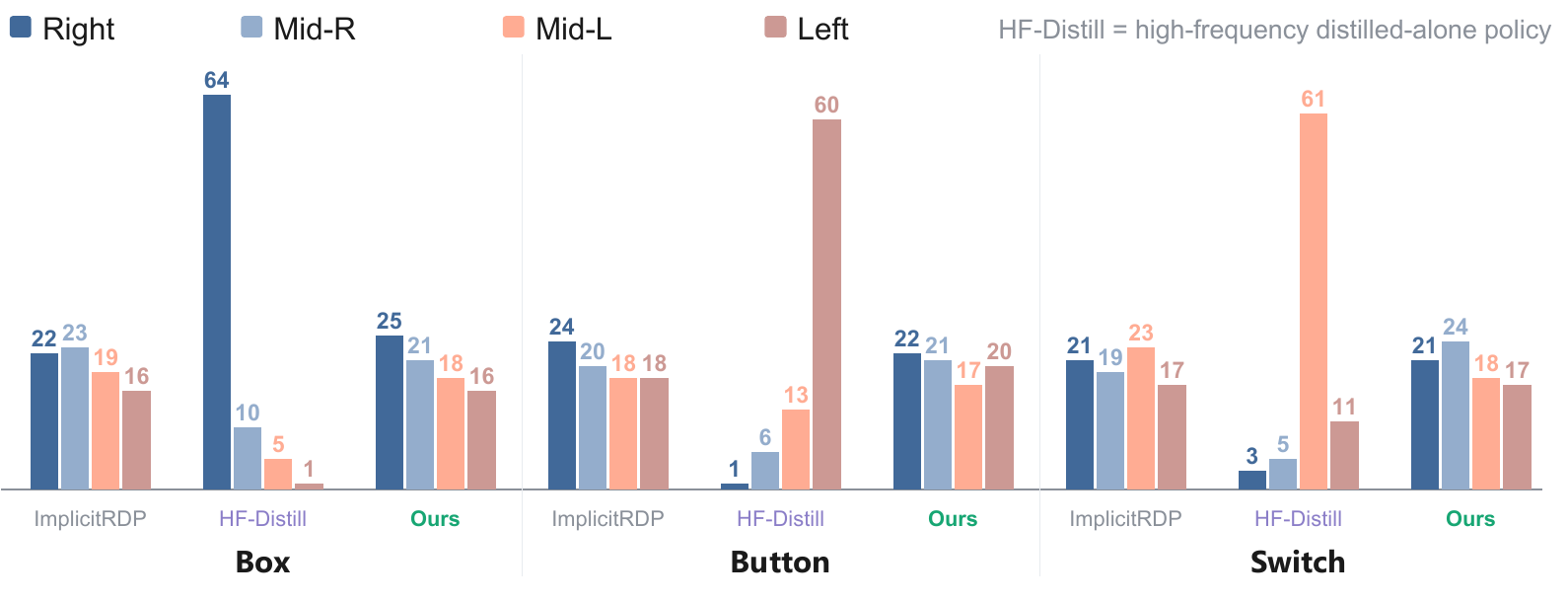}
    \caption{Pre-contact mode distributions across the three tasks. The
    high-frequency distilled alone policy collapses to one dominant approach mode,
    whereas FA-RDP and ImplicitRDP preserve multimodal distribution over
    pre-contact approaches.}
    \label{fig:multimodal_mode_distribution}
    \end{figure}

\subsubsection{Multimodal Preservation (Q2)}
We count the observed pre-contact trajectories in the right, middle-right,
middle-left, and left modes; the resulting distributions are visualized in
Fig.~\ref{fig:multimodal_mode_distribution}. For the multimodality study, we
run 80 trials per task; ImplicitRDP and FA-RDP cover all four modes, whereas the
high-frequency distilled alone policy concentrates in one mode.
This indicates that FA-RDP preserves the multimodal capability of multi-step
diffusion before contact while enabling high-frequency force response after
contact.

\begin{figure}[H]
    \centering
    \includegraphics[width=0.78\columnwidth]{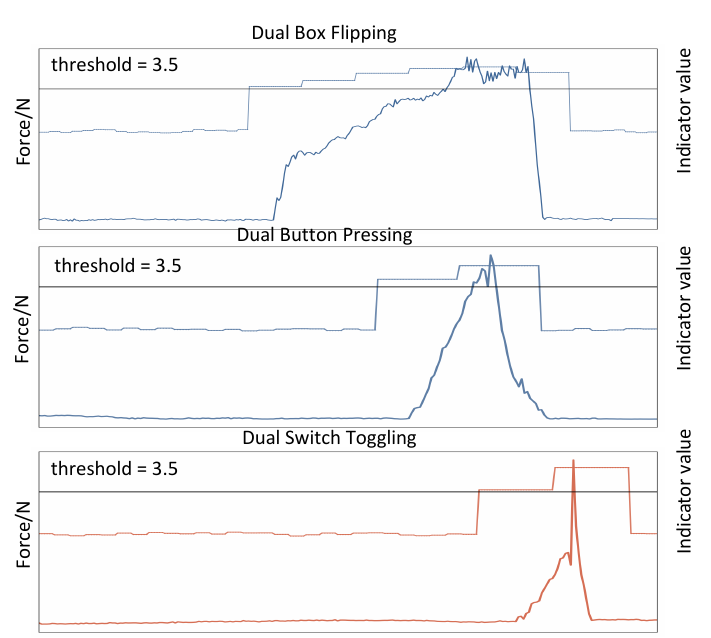}
    \caption{Multimodality indicator value and force curves on the three contact-rich
    tasks. The indicator value remains lower before contact and during target approach, then
    rises with the force response in contact-rich phases, supporting the switch from
    the low-frequency sampler to the high-frequency distilled sampler.}
    \label{fig:multimodality_force_curve}
    \end{figure}
\subsubsection{Indicator Accuracy (Q3)}
Q3 evaluates whether the multimodality indicator improves performance by
switching between the low-frequency sampler and the high-frequency
distilled sampler. We compare FA-RDP with the high-frequency distilled alone
policy in Table~\ref{tab:indicator_switching}. The latter always uses the
distilled high-frequency policy, whereas FA-RDP uses the low-frequency sampler
before contact and during target approach, then switches to the high-frequency
distilled sampler when contact-rich feedback becomes informative.

\begin{table}[H]
\caption{Indicator-guided switching comparison. Box, Button, and Switch denote
the three tasks.}
\label{tab:indicator_switching}
\centering
\scriptsize
\setlength{\tabcolsep}{1pt}
\renewcommand{\arraystretch}{1.08}
\begin{tabular*}{\columnwidth}{@{\extracolsep{\fill}}lcccc@{}}
\hline
\textbf{Method} & \textbf{Box} & \textbf{Button} & \textbf{Switch} & \textbf{Avg.} \\
\hline
High-frequency distilled alone policy & 12/20 & 14/20 & 11/20 & 61.7\% \\
\textbf{FA-RDP (Ours)} & \textbf{14/20} & \textbf{18/20} & \textbf{17/20} & \textbf{81.7\%} \\
\hline
\end{tabular*}
\end{table}

The multimodality and force curves in
Fig.~\ref{fig:multimodality_force_curve} show the temporal behavior of the
indicator. The indicator value stays low before contact, when multiple collision-free
approaches remain valid, so FA-RDP keeps the low-frequency diffusion sampler
active for smooth target approach before contact. The indicator rises
with the force response after contact, when the robot must maintain contact and
limit the applied force, so FA-RDP selects the high-frequency distilled sampler
for reactive execution. Compared with the high-frequency distilled alone
policy, this indicator-guided switching improves the average success rate from
$61.7\%$ to $81.7\%$.

\subsubsection{Distillation Method Comparison (Q4)}
Q4 evaluates whether our distillation performs better than other distillation
methods. All methods in
Table~\ref{tab:distillation_ablation} use the high-frequency mode, and
multimodality-based selection is not used. The high-frequency alone policy is
the original policy before distillation, and the other rows use the one-step
objectives defined in Sec.~\ref{subsec:baselines}.

\begin{table}[H]
\caption{High-frequency distillation comparison. Box, Button, and Switch denote
the three tasks.}
\label{tab:distillation_ablation}
\centering
\scriptsize
\setlength{\tabcolsep}{1pt}
\renewcommand{\arraystretch}{1.08}
\begin{tabular*}{\columnwidth}{@{\extracolsep{\fill}}lcccc@{}}
\hline
\textbf{Method} & \textbf{Box} & \textbf{Button} & \textbf{Switch} & \textbf{Avg.} \\
\hline
High-frequency alone policy & 4/20 & 3/20 & 5/20 & 20.0\% \\
MeanFlow Policy & 0/20 & 0/20 & 1/20 & 1.7\% \\
Consistency Policy & 0/20 & 0/20 & 1/20 & 1.7\% \\
\textbf{High-frequency distilled alone policy} & \textbf{12/20} & \textbf{14/20} & \textbf{11/20} & \textbf{61.7\%} \\
\hline
\end{tabular*}
\end{table}

Without distillation, running the original multi-step high-frequency policy
with one-step sampling results in insufficient prediction accuracy, causing
performance degradation. MeanFlow Policy and
Consistency Policy each achieve only $1.7\%$ average success, whereas our
high-frequency distilled policy reaches $61.7\%$. These results indicate
the effectiveness of manifold consistency distillation with SRL regularization
for one-step high-frequency control.

\section{Conclusion}

We presented FA-RDP, a frequency-adaptive reactive diffusion
policy that addresses the phase-dependent tradeoff between pre-contact
diversity and post-contact reactivity. Its shared multi-frequency visual-force
Transformer predicts multi-frequency action chunks, the multimodality indicator
guides frequency selection, and manifold consistency distillation provides
one-step high-frequency prediction on the robot action manifold. Across the three tasks,
FA-RDP combines broad pre-contact mode coverage with the highest average success rate
of $81.7\%$, demonstrating the effectiveness of our method. However, the current
evaluation is limited to visual-force input without testing other sensing modalities,
uses single-task policies rather than multi-task learning, and requires a three-stage
training procedure.

\bibliographystyle{IEEEtran}
\bibliography{references}
\end{document}